# Unmasking Removal-Budget Confounding: A Matched Operating-Point Evaluation Framework for Adaptive Data Cleaning

WEI-HSIANG CHEN
Department of Electrical Engineering, National Taiwan Ocean Univ., Keelung City, Taiwan
chenw4404@gmail.com

PIN-HSUAN YU
Department of Electrical Engineering, National Taiwan Ocean Univ., Keelung City, Taiwan
11453055@mail.ntou.edu.tw

CHEN-HSUAN FANG
Department of Electrical Engineering, National Taiwan Ocean Univ., Keelung City, Taiwan
samfang1108@gmail.com

JUNG-HUA WANG*
AI Research Center, National Taiwan Ocean Univ., Keelung City, Taiwan.
jhwang@email.ntou.edu.tw

***Abstract***—Adaptive data-cleaning methods replace manual filtering thresholds with data-driven partitions. However, changing the partition granularity, the number of groups used to segment samples by estimated corruption risk, can implicitly shift the decision boundary and alter the overall number of removed samples. This creates a bias known as *removal-budget confounding*, where apparent gains in metrics like precision or false-positive rate reflect a smaller removal budget rather than superior corruption discrimination. To address this evaluation bias, we introduce an operating-point-aware evaluation framework that evaluates methods using matched-budget and matched-recall controls alongside threshold-independent metrics (AUROC and AUPRC). We test this framework on a multi-cue adaptive cleaner redesign featuring a reweighted learning-difficulty cue, an auxiliary Euclidean-distance cue, and increased partition granularity intended to isolate clean-but-difficult samples. While naive evaluations (assessing configurations at their own induced operating points) suggest substantial performance improvements for the redesign, these gains disappear once operating points are equalized. False-positive decomposition reveals that clean-but-difficult samples primarily drive error counts at low corruption rates, become threshold-dependent at moderate corruption, and contribute negligibly under severe corruption. Experiments on CIFAR-10 and ImageNet-100 demonstrate that most performance differences observed in naive evaluation shrink or vanish at low-to-moderate corruption when operating points are matched. True ranking advantages only remain in specific low-prevalence settings and in high-recall regions under severe corruption. These findings highlight that adaptive cleaning methods must be benchmarked at matched operating points to ensure performance gains reflect genuine corruption discrimination.

## 1 INTRODUCTION

Data cleaning serves as a vital foundation for real-world AI applications, where system reliability directly hinges on the integrity of underlying training sets. In domain-specific fields such as medicine, where clinical decision-making relies on high-fidelity diagnostic imaging [1], and aquaculture, where anomalous fish behaviors detection [2] and species tracking face severe underwater visual noise and sensor distortions, unaddressed data corruption can catastrophically degrade model performance. Systemic label noise, structural outliers, and feature-level anomalies distort decision boundaries, leading to poor generalization and unreliable risk assessments in high-stakes environments. Robust data-cleaning frameworks are therefore essential to ensure that models learn from true underlying patterns rather than spurious noise.

Adaptive image data-cleaning techniques [3,4,5,6] increasingly swap static, manually set filtering thresholds for data-driven partitions built on sample-reliability scores. While score-based partitioning removes one form of researcher discretion, it fails to eliminate the operating point—the fundamental balance between total samples removed and corruption detected. A key driver of this boundary is partition granularity, defined as the number of risk

*Correspondent author

groups used to divide data [5]. Adjusting this granularity shifts the removal cutoff, directly changing the overall removal count.

Contemporary vision datasets suffer from diverse corruptions, ranging from mislabeled annotations to feature anomalies driven by sensor noise, occlusion, or distribution shifts. Multi-cue cleaning frameworks tackle this multi-faceted noise by combining metrics for local neighborhood agreement, global structural alignment, and model learning dynamics to partition data into distinct risk categories [7]. Samples in low-risk partitions are kept, while high-risk groups are purged. While finer partitioning promises better isolation of complex noise modes or clean-but-challenging instances, it inadvertently alters the removal volume, muddying performance comparisons under the native assessment.

This raises a crucial question: does a configuration reporting fewer false positives or higher precision truly discriminate corruption better, or is it simply removing fewer samples? Native evaluation assesses each configuration at its own induced operating point, rendering it incapable of separating true detection capability from shifting removal counts or recall. To resolve this ambiguity, we compare configurations under matched removal budgets, matched corruption recall, and threshold-independent AUROC/AUPRC. The matched controls equalize operating points, while AUROC and AUPRC measure ranking quality—how effectively risk scores order corrupted samples ahead of genuinely clean ones—which we refer to as discrimination throughout the paper.

At low-to-moderate corruption rates, performance gaps across partition granularities largely vanish once operating points are matched, though threshold-independent metrics show that ranking quality can still differ when corruption is rare. At the highest corruption rates, a genuine difference survives matching and is concentrated in the high-recall region. Furthermore, a redesign that reweights learning difficulty, adds an auxiliary Euclidean-distance cue, and uses one additional component appears better under native evaluation but provides no advantage under matched controls. False-positive decomposition reveals that clean-but-difficult samples—genuinely clean examples that are hard to learn—account for errors mainly at low corruption, and that their estimated contribution depends heavily on the chosen difficulty threshold.

These findings lead to three main contributions. (1) **Identifying Granularity Bias:** We establish partition granularity as an implicit determinant of removal budget. (2) **Controling Evaluation Framework**: We provide matched-budget, matched-recall, and threshold-independent evaluations that separate operating-point effects from ranking quality. (3) **Exposing Confounding Dynamics:** Through a controlled redesign and false-positive decomposition, we show how native comparisons can misattribute a smaller removal budget to improved detection.

We instantiate the analysis with a multi-cue mixture-based cleaner on two image datasets, but the evaluation principle applies whenever adaptive cleaning configurations induce different removal rates.

## 2 RELATED WORK

Adaptive data-cleaning methods have been developed from several complementary perspectives. Learning-dynamics approaches exploit the observation that clean samples are typically learned earlier and more consistently than corrupted ones, using measures such as prediction-margin accumulation [8] and per-sample loss trajectories [9]. Although effective, these scalar indicators may assign similarly high uncertainty to mislabeled samples and to rare but correctly labeled instances. Robust data-pruning methods instead optimize the subset of retained samples to maximize downstream re-labeling accuracy [10]. Geometric approaches identify corrupted samples through inconsistencies in local neighborhoods or global feature structure [11], but their effectiveness depends on neighborhood design and may deteriorate when corrupted samples distort the underlying representation. Other methods estimate label-error probabilities directly from model predictions without retraining [12] or identify noisy samples through repeated cross-validation [13], offering greater model independence at the expense of higher computational cost.

Our previous work [7] combines local, global, and learning-dynamics cues within an adaptive mixture-model framework, replacing manually selected filtering thresholds with data-driven partitioning. However, adaptive partitioning introduces a less recognized challenge: modifying partition granularity or cue design implicitly changes the removal budget, causing different configurations to operate at different decision boundaries. Consequently, native comparisons may reflect differences in removal rates rather than genuine corruption discrimination. This issue is closely related to recent benchmarking studies that standardize the number of flagged samples to enable fair comparison among noisy-label detectors [14]. Building on this principle, we evaluate adaptive cleaners using matched-budget and matched-recall controls while complementing them with threshold-independent ranking metrics to disentangle operating-point effects from discrimination quality.

The proposed evaluation framework is further motivated by two related research directions. In model-based clustering, the number of mixture components is commonly selected using criteria such as BIC [15], yet the statistically preferred model may still correspond to a different operating point, making downstream comparisons inherently unequal. Likewise, anomaly-detection algorithms such as Isolation Forest explicitly specify an expected contamination rate [16], whereas adaptive partitioning determines an equivalent removal rate implicitly through its model structure or retention policy. Finally, our analysis of clean-but-difficult samples is related to training-dynamics studies such as Dataset Cartography [17], which distinguish hard-to-learn from easy-to-learn examples. Unlike Dataset Cartography, where difficult samples may include mislabeled instances, our hard-clean subset contains only samples verified to be genuinely clean under the experimental corruption protocol. Collectively, these studies highlight the need to compare adaptive data-cleaning methods under controlled operating points rather than relying solely on their native partitions.

## **3** METHODOLOGY

### **3.1** Multi-cue reliability modeling and adaptive partitioning

A single reliability cue is insufficient to characterize the diverse forms of data corruption encountered in image datasets. Mislabeled samples may remain visually consistent with their semantic class while conflicting with local neighborhood or global class structure, whereas feature-level anomalies can retain correct labels yet remain persistently difficult for the model to learn. To capture these complementary characteristics, each sample is represented using three reliability cues: local neighborhood disagreement, global structural deviation, and learning difficulty. Rather than relying on any single indicator, their joint representation enables different corruption patterns to be distinguished even when individual cues are only weakly expressed.

Following [7], the evaluated cleaner consists of five sequential stages. A backbone network is first trained on the unfiltered dataset, after which feature embeddings are extracted for all samples. Local, global, and learning-difficulty cues are then computed and jointly standardized before adaptive partitioning. In the present implementation, partitioning is performed using a full-covariance Gaussian Mixture Model (GMM) optimized by the expectation-maximization algorithm [18]. The resulting mixture components are subsequently mapped to retained and removed groups according to the evaluation policy described in the following sections.

The proposed evaluation framework is not tied to the GMM itself. Here, the GMM serves only as a representative adaptive partitioning mechanism in which full covariance matrices capture interactions among the reliability cues and posterior component probabilities provide a continuous estimate of sample risk. Because the matched-budget and matched-recall evaluations introduced in Section 3.2 require only an ordered sample-risk ranking, the same protocol can be readily extended to alternative adaptive data-cleaning methods that produce comparable reliability scores.

Each sample is characterized by three complementary reliability cues. The first is a local neighborhood disagreement cue, which quantifies the consistency between a sample and its surrounding neighbors in the embedding space. For a sample $x_i$ with observed label $y_i$, the neighborhood size is determined in a class-conditioned manner as

$$K_{y_i} = \lfloor \sqrt{N_{y_i}} \rfloor \quad (1)$$

where $N_{y_i}$ denotes the number of training samples with observed label $y_i$. Defining the neighborhood size relative to the class population yields a more stable local representation than adopting a single fixed neighborhood across all classes. The local disagreement score is then computed as the proportion of the $K_{y_i}$ nearest neighbors of sample xi whose observed labels differ from *yi*:

$$r_i^{\text{Loc}} = \frac{1}{K_{y_i}} \sum_{j \in N_{K_{y_i}}(i)} 1(y_j \neq y_i) \quad (2)$$

A high local disagreement score indicates that the sample is embedded within a neighborhood dominated by samples from different observed classes, suggesting a higher likelihood of label corruption.

The global structural cue, $d_{1,i}$, is defined as the cosine distance between the embedding of sample $x_i$ and the centroid of its assigned k-means cluster, where the number of clusters equals the number of classes. The notation $d_1$ distinguishes this measure from the second distance metric, $d_2$ introduced later on. Whereas the local cue captures consistency within an immediate neighborhood, $d_1$ evaluates conformity to the overall class structure, enabling the detection of small, locally coherent groups of mislabeled samples.

The learning-difficulty cue identifies samples that remain difficult for the model to fit throughout training, including feature-level outliers that may nevertheless retain correct labels. Following the EL2N formulation [9], the raw difficulty score ei is computed as the average distance between the model's predicted class-probability vector and the one-hot encoding of the observed label for sample xi over the training process. To reduce the influence of class-specific variation, the raw scores are subsequently standardized within each observed class using the median and median absolute deviation (MAD):

$$s_i = \left(\frac{\text{MAD}(E_{y_i})}{0.6745}\right)^{-1} \{e_i - \text{median}(E_{y_i})\} \quad (3)$$

where $E_c$ = {$e_j$ | $y_j$ = c}. This normalization rescales the learning-difficulty distribution within each observed class without clipping or truncating extreme values. Compared with mean-based standardization, the median and MAD reduce the influence of atypical samples while preserving large difficulty values that may reflect corrupted data.

Before adaptive partitioning, all three reliability cues are standardized across the entire dataset. The partitioner is implemented as a full-covariance GMM. For evaluation, a genuinely clean sample refers to one unaffected by the experimental corruption protocol, whereas an estimated reliable sample denotes one assigned a low corruption risk by the cleaner. When ground-truth labels are available, the mixture component containing the highest proportion of genuinely clean samples is designated the clean-dominant component, with ties resolved by the absolute number of clean samples. Importantly, ground truth is used only to identify the retained component after model fitting and plays no role in estimating the mixture parameters. In practical deployments where ground truth is unavailable, component selection must instead rely on an unsupervised retention strategy, such as retaining the lowest-risk component. The implications of this assumption are discussed in Section 5.

### 3.2 Matched-budget and matched-recall controls

A false positive (FP) is defined as a genuinely clean sample assigned to a removed component, whereas a true negative (TN) is a genuinely clean sample assigned to a retained component. We analyze two-, three-, and four-component GMM configurations under corruption rates of 5%, 10%, 20%, and 40%. For each setting, FP samples are compared with both genuinely corrupted samples and TN samples in the original reliability-cue space and in a two-dimensional embedding projection to characterize their distribution and separability. Representative results are presented in Section 4.

Adaptive partitioning can shift the decision boundary and thereby change the number of removed samples. As a result, comparing false-positive counts or precision across configurations with different removal budgets is inherently confounded, since a more conservative configuration may achieve fewer FPs without providing better corruption discrimination. To isolate these effects, we employ two complementary evaluation controls. Each configuration first generates a continuous sample-risk score; for the GMM implementation, this score is defined as one minus the posterior probability of belonging to the retained component(s). The reference configuration specifies either the target removal budget or the target corruption recall. Under matched-budget evaluation, both configurations remove the same number of highest-risk samples according to their respective risk rankings. Under matched-recall evaluation, the decision threshold is adjusted until the compared configuration achieves the same corruption recall as the reference configuration. Because matched-recall evaluation requires ground-truth corruption labels, it serves solely as a diagnostic evaluation protocol.

As will be seen in Section 4, where we further decompose the native false-positive gap between two different *K*-configurations into removal-budget and evaluate risk ranking. Let $FP_K^{\mathrm{native}}$ denote the native FP count of the *K*-component configuration, and let $FP_K(B^*)$ denote its FP count when its risk ranking is evaluated at the reference budget $B^*$, which is the native removal count of the lower-*K* reference configuration. The native gap $\Delta_{\mathrm{native}} = FP_{\mathrm{ref}}^{\mathrm{native}} - FP_K^{\mathrm{native}}$ is decomposed into a ranking component $\Delta_{\mathrm{ranking}} = FP_{\mathrm{ref}}^{\mathrm{native}} - FP_K(B^*)$, which remains after matching the removal budget, and a budget component $\Delta_{\mathrm{budget}} = \Delta_{\mathrm{native}} - \Delta_{\mathrm{ranking}}$. We report $100\times \Delta_{\mathrm{budget}} / \Delta_{\mathrm{native}}$ and $100\times\Delta_{\mathrm{ranking}}/\Delta_{\mathrm{native}}$. Because $\Delta_{\mathrm{ranking}}$ may be negative, the budget component can exceed 100% at 40% corruption rate, as will be justified in Table 3.

### 3.3 Stress-testing a difficulty-weighted redesign

In our work we found that many FPs are geometrically ambiguous rather than members of a distinct population. We therefore introduce an auxiliary Euclidean-distance cue $d_{2,i}$, defined as the distance between sample i's embedding and its assigned centroid, and combine it with the original difficulty score $s_i$ to obtain the distance-weighted score

$$s_i' = d_{2,i} \cdot s_i \tag{4}$$

The weighted score is standardized in the same way as the original difficulty cue. We evaluate three- and four-component GMM, where the additional component tests whether clean-but-difficult samples form a separable group. For four-component settings, the primary policy retains the two lowest-risk components. A secondary policy is also evaluated for the independent-cues variant to measure sensitivity to component interpretation. Finally, we test a variant that keeps Euclidean distance and learning difficulty as separate inputs rather than multiplying them. All comparisons use matched-budget and matched-recall controls across five random initializations.

### 3.4 Decomposing FPs by difficulty and geometry

To operationalize the clean-but-difficult hypothesis, we define hard-clean samples as genuinely clean samples ranked within the top X% of raw learning difficulty in their observed class, with X ∈ {10, 20, 30}. The remaining genuinely clean samples are labeled easy-clean. We then identify a stricter subset of hard-clean samples whose auxiliary Euclidean distance $d_{2,i}$ is no greater than the class median; these samples are difficult to learn but not unusually distant from their class structure. Because the hard-clean definition is derived from a score related to the cleaner's own difficulty cue, the analysis measures enrichment, that is, over-representation among FPs, rather than establishing a causal mechanism.

## **4** EXPERIMENTAL RESULTS

### 4.1 Experimental Setup

We use CIFAR-10 as the primary experimental dataset and ImageNet-100 as a cross-dataset replication. Both datasets follow the same mixed-corruption protocol. At each target corruption rate (5%, 10%, 20%, or 40%), 90% of corrupted samples receive synthetic symmetric label noise and the remaining 10% are replaced by composite visual outliers. Feature embeddings are obtained from a fully fine-tuned ResNet-50 trained for five epochs with AdamW (learning rate 3×10−4, weight decay 10−2), batch size 128, and channel normalization only. All experiments use PyTorch 2.7.1 with CUDA 11.8 on a single RTX 2080 GPU. The local cue uses cosine-similarity nearest-neighbor search, while the global cues use k-means with the number of clusters set to the number of classes.

The GMM uses full covariance matrices, covariance regularization, and 10 internal initializations. Native partition-granularity and normalization comparisons use one fixed outer initialization, whereas the central redesign stress test uses five. Before comparative analysis, we verified exact reproduction of a pre-existing baseline partition. In the four-component audit, the baseline retains one clean-dominant component; the redesign stress test retains two components to represent a possible clean-but-difficult population.

### 4.2 Analysis the effect of finer partition granularity

Table 1 summarizes native evaluation, meaning that each configuration is assessed at its own induced operating point. For the three-component baseline, the FPR is highest at 5% corruption and remains relatively stable from 10% to 40%, indicating that the low precision at 5% reflects an elevated false-alarm rate rather than only the low prevalence of corruption. Native FPR also increases with $K$ at every tested rate. At 5% corruption, it rises from 10.29% with two components to 29.71% with three and 32.78% with four. The corresponding FP samples occupy a broad region of the embedding space that overlaps both retained clean samples and genuinely corrupted samples, particularly label-noise cases, rather than forming a clearly separable cluster.

Matched-budget and matched-recall controls substantially change this interpretation at low-to-moderate corruption. At 5%, 10%, and 20%, the native differences among two-, three-, and four-component configurations largely disappear after the removal budget or recall target is matched. For example, when the three- and four-component rankings are evaluated at the two-component configuration's native removal budget at 5% corruption, precision becomes 33.5%~33.6%, essentially identical to the two-component native value of 33.5%. These results indicate that most native differences in this range arise from different operating points rather than different ranking quality.

The pattern changes at 40% corruption. Matching the three-component native recall of 97.97% requires the two-component ranking to remove about 87% of the evaluation set, producing an FPR of 80.3%, whereas the four-component ranking reaches the same recall at an FPR of 14.3%. Thus, the two-component ranking genuinely deteriorates in the high-recall region at the highest corruption rate tested. Removal-budget confounding therefore explains most low-to-moderate-rate differences, but it does not account for the severe-corruption ranking gap.

### 4.3 Renormalizing the difficulty cue: a native advantage reversed by matched recall

We compare five normalization configurations: the dataset-wide baseline, a geometric-only variant that omits the difficulty cue, and three class-wise variants that differ in how robust difficulty standardization is applied. Under native evaluation, the geometric-only variant appears best at low-to-moderate corruption. This apparent advantage disappears under matched-budget evaluation, where all five configurations converge to within approximately 5% of one another. Under matched recall, the ordering reverses: matching the other configurations' recall increases the geometric-only variant's FP count by about 3.2~3.3× at 5% corruption, 2.6~2.7× at 10%, and 1.26~1.27× at 20%, while the difference falls to about 1% at 40%. The native comparison therefore ranks the geometric-only variant first over most of the tested range, whereas the matched-recall comparison shows the opposite. The three class-wise variants remain close to the baseline and to one another under both controls.

### 4.4 Vanishing reduction in FP under matched controls

Table 2 compares performance of the difficulty-weighted redesign across four corruption rates and five random initializations when performing native evaluation on CIFAR-10. Because the number of clean samples decreases from about 47,500 at 5% corruption to 30,000 at 40%, the mean FP count is descriptive rather than directly comparable across rates. Macro-FPR, which gives equal weight to the FPR at each corruption rate, is the more appropriate normalized summary.

Native evaluation suggests a strong advantage for four components: mean FPs fall to 28%–35% of the corresponding three-component values and precision increases by about 20 percentage points. However, the baseline itself reaches 2,054 mean FPs and 72.7% precision with four components, matching or outperforming both redesigned variants. The apparent improvement therefore cannot be attributed to the new cue design alone.

Table 1: Native CIFAR-10 detection performance across *K* values.

| Ratio | *K* | Precision (%) | Recall (%) | FPR (%) | Label-noise removal (%) | Outlier removal (%) |
|---|---|---|---|---|---|---|
| 5% | 2 | 33.48 | 98.44 | 10.29 | 99.47 | 89.20 |
| 5% | 3 | 15.02 | 99.80 | 29.71 | 99.96 | 98.40 |
| 5% | 4 | 13.82 | 99.84 | 32.78 | 100.00 | 98.40 |
| 10% | 2 | 56.89 | 97.60 | 8.22 | 98.60 | 88.60 |
| 10% | 3 | 46.05 | 98.74 | 12.86 | 99.13 | 95.20 |
| 10% | 4 | 27.61 | 99.68 | 29.04 | 99.69 | 99.60 |
| 20% | 2 | 76.65 | 97.99 | 7.46 | 98.99 | 89.00 |
| 20% | 3 | 64.68 | 99.22 | 13.55 | 99.56 | 96.20 |
| 20% | 4 | 46.19 | 99.77 | 29.06 | 99.84 | 99.10 |
| 40% | 2 | 96.65 | 90.58 | 2.09 | 94.29 | 57.15 |
| 40% | 3 | 82.21 | 97.97 | 14.13 | 98.57 | 92.55 |
| 40% | 4 | 76.85 | 98.82 | 19.84 | 99.16 | 95.70 |

Table 2: Performance comparison of various configurations when performing native evaluation on CIFAR-10

| Configuration | Components | Mean FP | Precision | Recall % | Outlier removal % | Macro-FPR % |
|---|---|---|---|---|---|---|
| Baseline difficulty cue | 3 | 7,389 | 51.98 | 98.93 | 95.59 | 17.58 |
| Baseline difficulty cue | 4 | 2,054 | 72.67 | 94.78 | 74.21 | 4.84 |
| Distance-weighted difficulty cue | 3 | 7,786 | 50.42 | 99.00 | 95.43 | 18.63 |
| Independent distance and difficulty cues | 3 | 6,215 | 52.45 | 98.67 | 93.48 | 15.57 |
| Distance-weighted difficulty cue | 4 | 2,289 | 71.13 | 95.04 | 75.70 | 5.33 |
| Independent distance and difficulty cues | 4 | 2,158 | 72.30 | 94.63 | 73.29 | 5.12 |

Table 3: Removal-budget and ranking decomposition of the native 3vs4-component FP gap for CIFAR-10.

| Corruption ratio | Budget effect (% of gap) | Ranking effect (% of gap) |
|---|---|---|
| 5% | 100.01 | -0.01 |
| 10% | 100.61 | -0.61 |
| 20% | 100.78 | -0.78 |
| 40% | 117.64 | -17.64 |

Table 3 decomposes the baseline's native three-versus-four-component FP gap. Across corruption rates, the removal-budget component explains approximately 100%–118% of the gap, while the ranking component is near zero or negative. In other words, the fourth component improves the native FP count by imposing a more conservative removal boundary, not by improving the risk ranking at the matched budget.

Fig.1 confirms this interpretation. With three components, the distance-weighted variant remains worse than the baseline under both native and matched-budget evaluation. With four components, all three configurations converge to approximately 7,558–7,564 FPs when evaluated at the baseline removal budget, despite their much lower native FP counts. Matched-recall evaluation leads to the same conclusion: each redesigned variant produces more FPs than the baseline after recall is equalized, and the gap is larger with four components. Across five GMM initializations, the native and matched values vary by only a few FPs under fixed embeddings and corruption. This stability concerns repeated GMM fitting, not the full training and corruption-generation pipeline. The two retention policies for the four-component independent-cues variant agree at low-to-moderate corruption but diverge sharply at 40%, indicating that component interpretation is itself consequential.

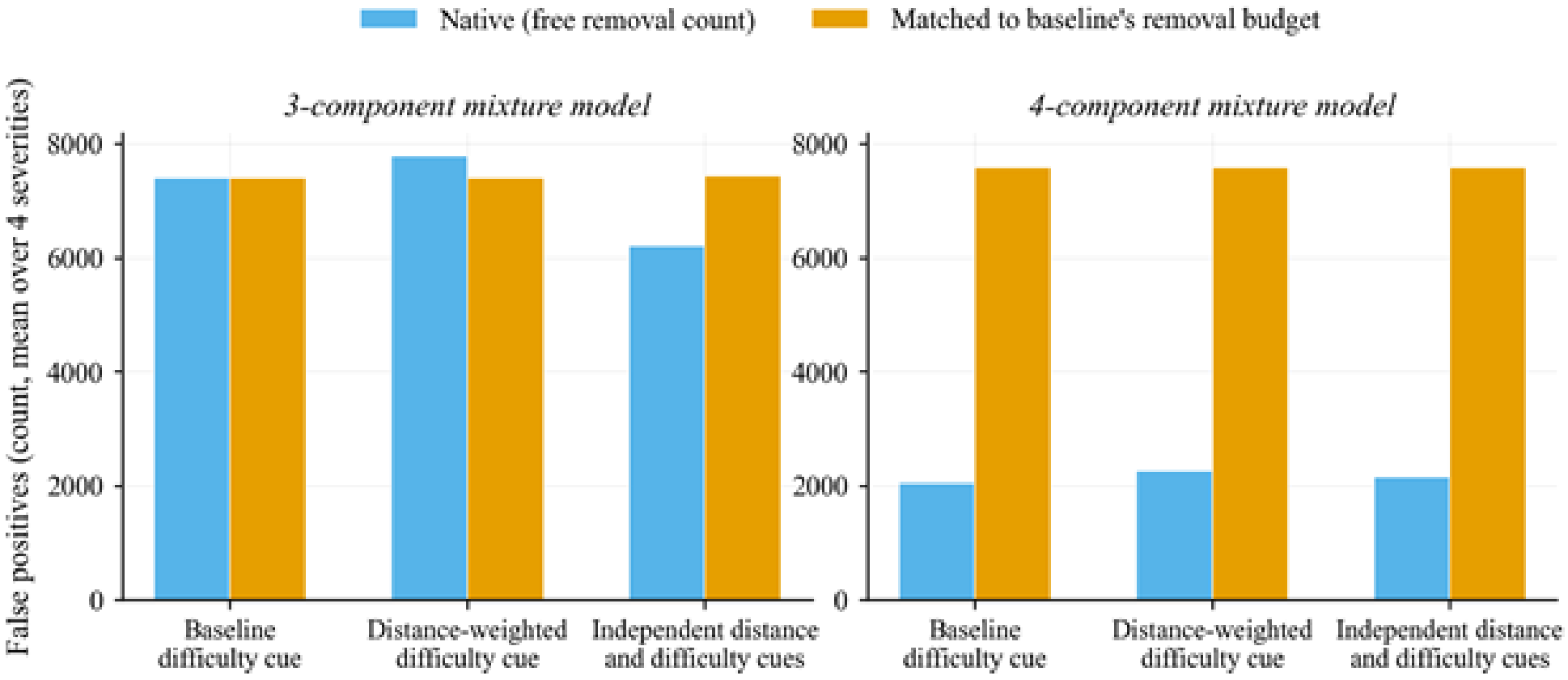

Figure 1: Native and matched-budget mean false-positive counts for three cue configurations under three- and four-component partitions on CIFAR-10.

### 4.5 False positives or clean-but-difficult?

Using the top-20% difficulty threshold, hard-clean samples account for 53.3% of FPs at 5% corruption and 73.8% at 10%, but only 12.0% at 20% and 0.12% at 40%. Their contribution therefore falls sharply as corruption becomes more prevalent. The stricter subset that is both hard-clean and no farther from its centroid than the class median accounts for at most 14.6% of FPs at any rate. Because the subset is defined using a score related to the cleaner's own difficulty cue, this result demonstrates enrichment rather than independent causal evidence.

The estimated contribution is also sensitive to the threshold used to define hard-clean samples. At 5% corruption, the FP share ranges from 18.2% to 81.3% when the threshold changes from the top 10% to the top 30%. At 20% corruption, the share reaches 65.2% under the most permissive definition, whereas at 40% it remains at or below 1.2% for all three thresholds. The clean-but-difficult hypothesis is therefore strongly supported only at low corruption, becomes definition-dependent at moderate corruption, and receives little support at 40%.

To assess whether removal-budget confounding generalizes beyond CIFAR-10, we replicate the native and matched-control evaluations on ImageNet-100 using the same corruption protocol. Owing to GPU memory constraints, feature embeddings at 20% corruption are extracted with a batch size of 64 instead of 128. A validation on a held-out set shows that the resulting embeddings are nearly identical (mean cosine similarity: 0.999999; mean relative L2 difference: 0.03%), suggesting that the reduced batch size is unlikely to materially affect the embedding representation. However, the complete downstream pipeline was not rerun with a batch size of 128.

ImageNet-100 exhibits the same overall behavior as CIFAR-10: native performance differences across partition granularities are substantially reduced once removal budgets are matched. Likewise, matched-recall differences remain small at 5% and 10% corruption but become considerably larger at 20% and 40%, indicating that ranking quality plays an increasingly important role as corruption becomes more severe. Unlike CIFAR-10, however, the transition from operating-point-dominated behavior to genuine ranking differences is more gradual, with comparable gaps observed at both 20% and 40% corruption. These findings demonstrate that removal-budget confounding is not specific to a particular dataset, while also indicating that the corruption level at which ranking differences become prominent is dataset dependent.

Table 4 shows threshold-independent AUROC and AUPRC for different $K$ values on both datasets. AUROC changes comparatively little, whereas AUPRC varies more strongly, particularly at lower corruption prevalence. This distinction refines the matched-control result: many differences at native operating points are driven by the removal budget, but the underlying global rankings are not identical.

Table 4: AUROC/AUPRC of the baseline configuration across *K* components and datasets.

| Dataset | Ratio | K=2, AUROC/AUPRC | K=3, AUROC/AUPRC | K=4, AUROC/AUPRC |
|---|---|---|---|---|
| CIFAR-10 | 5% | 0.98/0.57 | 0.96/0.41 | 0.96/0.38 |
| CIFAR-10 | 10% | 0.98/0.80 | 0.98/0.78 | 0.97/0.70 |
| CIFAR-10 | 20% | 0.98/0.87 | 0.98/0.86 | 0.97/0.83 |
| CIFAR-10 | 40% | 0.96/0.95 | 0.98/0.96 | 0.98/0.95 |
| ImageNet-100 | 5% | 0.93/0.28 | 0.92/0.26 | 0.89/0.20 |
| ImageNet-100 | 10% | 0.95/0.54 | 0.94/0.49 | 0.92/0.41 |
| ImageNet-100 | 20% | 0.98/0.95 | 0.96/0.75 | 0.95/0.70 |
| ImageNet-100 | 40% | 0.97/0.97 | 0.98/0.96 | 0.98/0.94 |

## **5** DISCUSSION

Across all experiments, partition granularity was found to influence both the native operating point and the underlying sample-risk ranking, and these two effects should therefore be evaluated independently. At low-to-moderate corruption rates, most performance differences observed under native evaluation disappeared once removal budgets or corruption recalls were matched, indicating that they were largely attributable to removal-budget confounding rather than improved corruption discrimination. At higher corruption rates, however, a genuine ranking advantage persisted in the high-recall region. The difficulty-weighted redesign further illustrates the importance of this distinction: although its lower native false-positive rate was primarily a consequence of a more conservative removal boundary, threshold-independent evaluation demonstrates that different partition granularities can still induce distinct global risk rankings, with meaningful differences becoming apparent under specific operating conditions.

The clean-but-difficult hypothesis likewise exhibits a strong dependence on corruption level. Hard-clean samples are substantially overrepresented among false positives at low corruption, but their estimated contribution becomes increasingly sensitive to the chosen difficulty threshold at moderate corruption and is negligible under 40% corruption. These findings further demonstrate that native performance alone cannot distinguish genuine discrimination improvements from operating-point effects, reinforcing the need to treat matched-budget and matched-recall evaluations as standard benchmarking protocols rather than optional diagnostic analyses. Although finer partitioning consistently outperforms the two-component configuration in the high-recall regime at severe corruption, exploratory covariance analysis does not explain the underlying mechanism. We therefore refrain from attributing this behavior to a specific cause and instead report it as an unresolved empirical observation.

A complementary FPR-at-recall analysis further localizes the ranking advantage observed under severe corruption. At 40% corruption, all *K*-component configurations achieve 90% recall with similarly low FPRs. However, beyond this operating point, the two-component configuration deteriorates rapidly: its FPR increases to 31.9% at 95% recall and 78.4% at 98% recall, whereas the corresponding values remain at 5.6% and approximately 14.3%–14.4% for the three- and four-component configurations. These results indicate that the principal benefit of increasing the partition granularity is realized when operating in the high-recall regime. Specifically, moving from two to three components substantially improves ranking stability, while introducing a fourth component yields only marginal additional gains. Figure 2 presents the corresponding precision–recall curves, providing a complementary view of this behavior.

Two exploratory end-to-end reruns with different random seeds reproduced some, but not all, pairwise performance rankings. Consequently, the current evidence provides stronger support for the stability of the adaptive partitioning stage under fixed feature embeddings than for the robustness of the complete training and corruption-generation pipeline. In the normalization ablation, each comparison uses a single outer configuration, whereas the five-seed stress test varies only the GMM initialization while keeping the feature embeddings and corruption realization fixed. Because each GMM fit already performs ten internal expectation-maximization initializations, these experiments primarily assess partition-fitting stability. By contrast, the two end-to-end reruns remain exploratory, employ a reduced training batch size, and are insufficient to establish full-pipeline robustness. More comprehensive validation will require additional independent backbone-training and corruption-generation seeds.

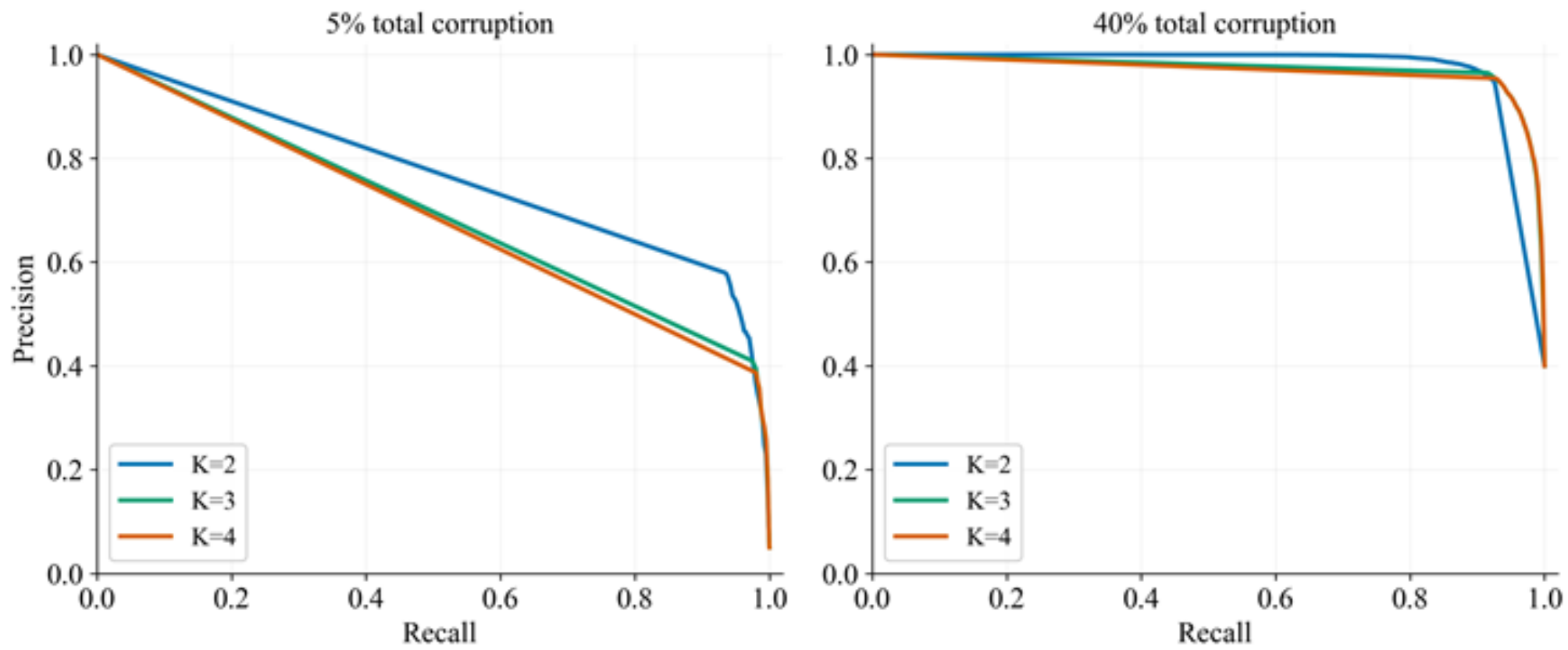

Figure 2: Precision-recall curves for the two-, three-, and four-component baseline on CIFAR-10 at 5% (left) and 40%(right) corruption.

The clean-but-difficult decomposition is inherently associational rather than causal because hard-clean samples are defined using a learning-difficulty measure derived from the cleaner itself. Establishing a causal relationship would require validation with an independent difficulty metric. Similarly, matched-recall evaluation and ground-truth-based component identification are diagnostic procedures that depend on access to ground-truth labels and are therefore unsuitable for deployment. Within the evaluated settings, an unsupervised lowest-risk retention rule produced the same retained component(s) as the oracle rule for the two- and three-component configurations up to 40% corruption, and likewise selected the same retained pair in the four-component baseline. However, this agreement has not been validated beyond 40% corruption, and alternative retention policies for the four-component configuration diverge substantially at higher corruption levels.

The cross-dataset experiments demonstrate that removal-budget confounding is not unique to CIFAR-10. ImageNet-100 reproduces the same qualitative confounding effect and similarly exhibits larger ranking differences under severe corruption, although the transition between operating-point effects and genuine ranking differences occurs less distinctly than on CIFAR-10. The feature-level batch-size comparison suggests that the difference observed at 20% corruption is unlikely to be attributable to the embedding extraction process, although a small downstream influence cannot be completely excluded without rerunning the full experimental pipeline. More broadly, the present study is limited to a single mixture-based cleaning framework, one backbone architecture, synthetic symmetric label noise, constructed visual outliers, and two image benchmarks. While the proposed evaluation protocol requires only a continuous sample-risk ranking and is therefore applicable to a wide range of adaptive data-cleaning methods, the prevalence and severity of removal-budget confounding under real-world conditions, including asymmetric and instance-dependent label noise, alternative cleaning paradigms, and non-image domains, remain important directions for future investigation.

## **6** CONCLUSION

Adaptive data-cleaning methods are commonly evaluated using their native operating points, making it difficult to distinguish genuine corruption discrimination from differences in removal budget. Through a mixture-based case study, this work demonstrates that partition granularity influences both the induced removal boundary and the underlying sample-risk ranking. At low-to-moderate corruption rates, most apparent performance differences disappear once removal budgets or corruption recalls are matched, indicating that they are primarily attributable to removal-budget confounding. Under severe corruption, however, a genuine ranking advantage persists in the high-recall region, highlighting that operating-point effects and ranking quality represent distinct aspects of cleaner performance. The proposed evaluation framework further reveals that the apparent improvements of the difficulty-weighted redesign and the normalization variants arise largely from changes in operating point rather than enhanced corruption discrimination. The same qualitative confounding phenomenon is consistently observed on both CIFAR-10 and ImageNet-100, although the corruption level at which ranking differences become pronounced remains dataset dependent. These findings demonstrate that native evaluation alone is insufficient for assessing adaptive data-cleaning methods. Future studies should therefore complement native results with matched-budget, matched-recall, and threshold-independent evaluations to separate operating-point effects from true discrimination capability and enable fair, interpretable comparisons across adaptive cleaning strategies.

**ACKNOWLEDGMENTS**

The work was partially supported by National Science Technology Council under Grant No.115-2221-E-019-055 and the Ministry of Agriculture (MOA) under Grant No.115AS-17.3.2-AS-0.